\documentclass[12pt]{iopart}
\expandafter\let\csname equation*\endcsname\relax
\expandafter\let\csname endequation*\endcsname\relax
\usepackage{amsmath,amssymb,amsfonts}
\usepackage{cite}
\usepackage{algorithmic}
\usepackage{textcomp}
\usepackage{xcolor}
\usepackage{hyperref}
\usepackage{graphicx} % Required for inserting images
\usepackage{makecell}
\usepackage{multirow}
\usepackage{bm}
\usepackage{verbatim}
\usepackage{caption}
\usepackage[commandnameprefix=ifneeded]{changes}
\usepackage{soul}

\begin{document}

\title[A Kernel Space-based Multidimensional Sparse Model for Dynamic PET Image Denoising]{A Kernel Space-based Multidimensional Sparse Model for Dynamic PET Image Denoising}
\author{Xiaodong Kuang$^{1}$, Bingxuan Li$^{2}$, Yuan Liu$^3$, Fan Rao$^4$, Gege Ma$^5$, Qingguo Xie$^2$, Greta S P Mok$^6$, Huafeng Liu$^7$ and Wentao Zhu$^{8,*}$}
\address{$^1$ Zhejiang Polytechnic University of Mechanical and Electrical Engineering, Hangzhou, China}
\address{$^2$ Institute of Artificial Intelligence, Hefei Comprehensive National Science Center, Hefei, China}
\address{$^3$ School of Electrical and Optical Engineering, Nanjing University of Science and Technology, Nanjing, China}
\address{$^4$ Zhejiang Lab, Hangzhou, China}
\address{$^5$ School of Intelligent Science and Technology, Hangzhou Institute for Advanced Study, University of Chinese Academy of Sciences, Hangzhou, China}
\address{$^6$ Biomedical Imaging Laboratory (BIG), Department of Electrical and Computer Engineering, Faculty of Science and Technology, University of Macau, China}
\address{$^7$ College of Optical Science and Engineering, Zhejiang University, Hangzhou, Zhejiang}
\address{$^8$ College of Biomedical Engineering and Instrument Science, Zhejiang University, Hangzhou 310027, China, and also with Zhejiang Lab, Hangzhou 311121, China}
\address{$^*$ Authors to whom any correspondence should be addressed. }
\ead{\url{zhuwentao.ee@gmail.com}}
\vspace{10pt}

\begin{abstract}
Achieving high image quality for temporal frames in dynamic positron emission tomography (PET) is challenging due to the limited statistic especially for the short frames. Recent studies have shown that deep learning (DL) is useful in a wide range of medical image denoising tasks. In this paper, we propose a model-based neural network for dynamic PET image denoising. The inter-frame spatial correlation and intra-frame structural consistency in dynamic PET are used to establish the kernel space-based multidimensional sparse (KMDS) model. 
We then substitute the inherent forms of the parameter estimation with neural networks to enable adaptive parameters optimization, forming the end-to-end neural KMDS-Net.
Extensive experimental results from simulated and real data demonstrate that the neural KMDS-Net exhibits strong denoising performance for dynamic PET, outperforming previous baseline methods. The proposed method may be used to effectively achieve high temporal and spatial resolution for dynamic PET. Our source code is available at \textit{https://github.com/Kuangxd/Neural-KMDS-Net/tree/main}.

\end{abstract}

\section{Introduction}
\label{sec:intro}
Positron emission tomography (PET) is widely used to evaluate the biological metabolism processes \textit{in vivo} \cite{ref1}. Due to various physical degradation factors and limited statistics, image reconstruction from PET data usually results in poor signal-to-noise ratio (SNR) and spatial resolution. Dynamic PET imaging may be more challenging when the counts become even lower for the early short frames when high temporal resolution is desired. Image denoising is important in order to acquired image frames with both high temporal and spatial resolution in dynamic PET.

Dynamic PET denoising methods may be divided into two categories, i.e., model-based and learning-based. The model-based methods include various denoising strategies within or after reconstruction. For reconstruction in dynamic PET, researchers have introduced various prior models for controlling image noise, such as edge preservation priors \cite{ref2}, voxel kinetics priors \cite{ref3}, anatomical priors \cite{ref4} and kernel models \cite{ref5}. Non-local means \cite{ref25} and highly constrained backProjection (HYPR) \cite{ref7, ref27} are used to further improve the kernel methods. On the other hand, post-reconstruction denoising methods such as nonlocal mean filtering (NLM) \cite{ref6}, block matching and 3D filtering (BM3D) \cite{ref8}, image guided filtering (IGF) \cite{ref9} and wavelet denoising \cite{ref10} have been proposed for dynamic PET. These model-based methods are physically interpretable. However, they heavily rely on the accuracy of system model, the choice of regularization terms, high-computational optimization and complex hyperparameter tuning.

The deep learning (DL)-based denoising methods have produced appealing results in computed tomography images \cite{ref44} and multi-modal medical images \cite{ref45} and have been successfully applied to dynamic PET. The convolutional neural network (CNN)-based methods such as population-based model \cite{ref20} and three deep learning enhancement models \cite{ref41} can effectively suppress noise yet may easily cause oversmoothed results. Generative Adversarial Networks (GANs) including dual GANs \cite{ref43}, conditional GANs \cite{ref11} and cycle GANs \cite{ref42} have been proposed to restore image details better, but the stability of training and the authenticity of generated details remain uncertain. Recently, denoising diffusion probabilistic model (DDPM) and score function-based model have been developed to learn the transition from a normal distribution to a specific data distribution, generating high-quality results that are virtually quite consistent with real data \cite{ref37}.
These methods directly learn the nonlinear mapping between the "noisy and clean" PET training data, ignoring the subjective physical mechanisms of PET imaging. Benefiting from their strong generalization ability, DL-based methods have demonstrated promising results when trained with a large amount of training data. However, their denoising performance may still be limited in complex clinical scenarios. To alleviate this issue, DL-based methods are incorporated into the image reconstruction process. For example, deep learning priors are used to accurately represent pixel features \cite{ref16} and coefficient images \cite{ref17} in kernel methods and also to improve image quality in iterative reconstruction \cite{ref24}. Although these methods have demonstrated strong generalization capabilities, the hyperparameter tuning and computational cost are usually challenging.

\begin{figure*}[t]%[th!b]
    \centering
        \includegraphics[width=1\textwidth]{"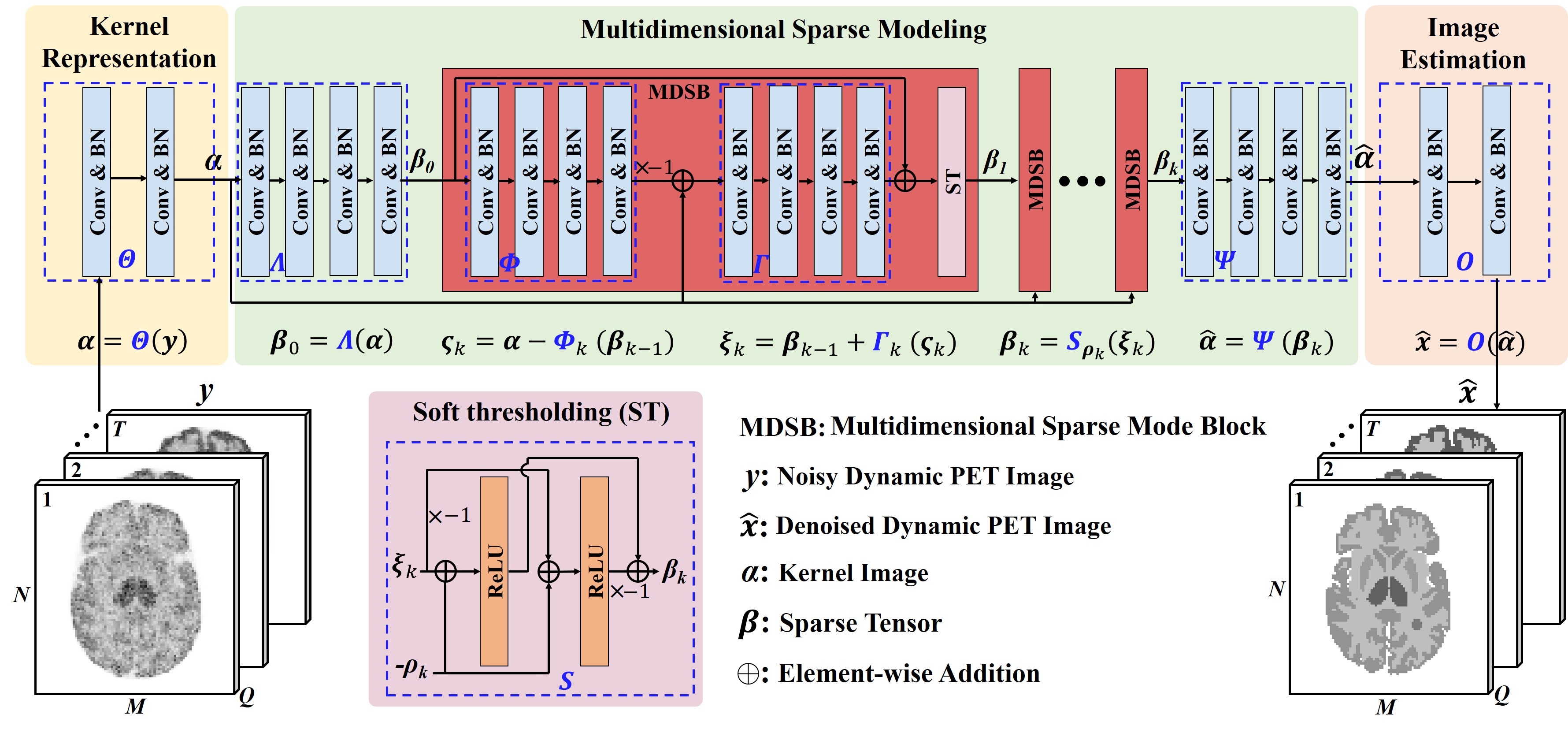"}
        % \caption{Illustration of the proposed method. First, the noisy PET image \(\textbf{\textit{y}}\) is converted into the kernel representation \(\boldsymbol\alpha\) using the neural network .}
        \caption{Illustration of the proposed method. First, the noisy PET image \(\textbf{\textit{y}}\) is converted into the kernel representation \(\boldsymbol{{\alpha}}\) using the neural network \(\boldsymbol{{\Theta}}\). The sparse representation \(\boldsymbol{{\beta}}\) is then approximated from \(\boldsymbol{ \alpha}\) using a series of multidimensional sparse coding blocks and used to compute the estimated kernel representation \(\hat{\boldsymbol{{\alpha}}}\). Finally, a neural network \textit{\textbf{O}} is used to estimate the denoised PET image \(\hat{\textbf{\textit{x}}}\).}
    \label{fig:1}
\end{figure*}

It is preferable to design a dynamic PET denoising method that combines the advantages of the model-based and DL-based approaches, i.e., with high interpretability and learnability.
Kernel methods have been proven to effectively reduce noise in dynamic PET images.
Its performance, however, relies heavily on the choice of hyperparameters, such as the kernel functions, composite images and search range in $k$-nearest neighbors ($k$NN). It has been reported that a trainable network can be used to automatically learn these hyperparameters in PET \cite{ref18} and hyperspectral \cite{ref36} image denoising. Moreover, the kernel methods usually take 3D images as input instead of 4D ones in previous literature, which does not fully exploit the spatial-temporal correlation. On the other hand, Qi \textit{et al.} proposed a $N$-order tensor which may be represented as a multidimensional signal to exploit the correlations in each dimension \cite{ref31}. Xiong \textit{et al.} applied this concept to construct a physical model in the image domain and used a network (SMDS-Net \cite{ref36}) to learn the hyperparameters. Inspired by these previous works, we aim at introducing the kernel methods in the image domain and use multidimensional sparse coding to establish the relationship between temporal frames. Therefore in this paper, a model-based neural network is proposed for dynamic PET image denoising which integrates both the temporal and spatial image prior information, as shown in Fig. \ref{fig:1}. Specifically, we build a kernel space-based multidimensional sparse (KMDS) model under the framework of multidimensional tensor sparse representation. The KMDS model has three steps. First, the noisy dynamic PET image is mapped into the kernel space to obtain spatial-temporal correlations, yielding a kernel image. Second, the KMDS model uses a multidimensional tensor sparse model which sparsely encodes the kernel image with multidimensional dictionaries to reduce image redundancy. Finally, a denoised PET image is obtained through a linear combination of local pixels in the encoded kernel image. We then develop an end-to-end deep neural network named neural KMDS-Net, which substitutes each module in the optimization algorithm of the KMDS model with convolutional neural networks. Neural KMDS-Net consists of three stages, i.e., kernel representation, multidimensional sparse modeling, and image estimation. Although both our method and SMDS-Net use the multi-dimensional sparse representation framework, the feature space where sparse tensors reside and the optimization methods for the parameters in this framework differ. For details, please refer to Section \ref{sec:dis} in the relevant section.

This paper is organized as follows. Section \ref{sec:method} describes the proposed neural KMDS-Net. We then present the simulation studies in Section \ref{sec:sim} and experimental results on real patient data in Section \ref{sec:real} to demonstrate the advantages of neural KMDS-Net over other state-of-the-art model-based and DL-based methods. Finally, discussions and conclusions are drawn in Section \ref{sec:dis} and \ref{sec:con}.

\section{Method}
\label{sec:method}
\subsection{KMDS model formulation}
Given a dynamic PET image frame \textit{\textbf{y}} with \(M\times N \times Q\) pixels spatially and \textit{T} frames, the observation model may be formulated as
\begin{equation}
    \textbf{\textit{y} = \textit{x} + \textit{n}}
\end{equation}
where \(\textbf{\textit{x}}\in R^{M\times N \times Q\times T}\) represents the true dynamic PET image to be estimated and \(\textbf{\textit{n}}\in R^{M\times N\times Q\times T}\) represents the additive noise. The image-domain additive noise model is adopted for the post-reconstruction denoising task, focusing on mitigating residual noise in the reconstructed image space rather than modeling the PET physical acquisition process. One common solution for the above problem is to introduce prior information as regularization and then optimize the corresponding cost function.

As stated in \cite{ref5}, the image \textbf{\textit{x}} can be represented by a kernel matrix \(\textbf{\textit{K}}\in R^{(MNQ)\times (MNQ)}\) and a coefficient image \(\boldsymbol{\alpha}\in R^{M\times N\times Q\times T}\)
\begin{equation}
    \tau(\textbf{\textit{x}})=\textbf{\textit{K}}\tau(\boldsymbol{\alpha})
\end{equation}
where \(\tau(\cdot)\)  represents the operation of transforming a matrix of size \(M\times N\times Q\times T\) into a matrix of size \(MNQ\times T\). \textbf{\textit{K}} is typically obtained using composite images \cite{ref5}. In this way, each pixel intensity in the image \textbf{\textit{x}} is expressed as a linear function in the kernel space. Next, the prior knowledge of the correlations between any of 2 frames is used to model \(\boldsymbol{\alpha}\).

For dynamic PET images, each frame \(\textbf{\textit{x}}_{t}\in R^{M\times N\times Q}\) has the same structure if motion does not occur. \textbf{\textit{K}} is a pre-computed matrix, implying that in the kernel space, each frame \(\boldsymbol{\alpha}_{t}\in R^{M\times N\times Q}\) shares structural similarity. To exploit the correlations in every frame, we model \(\boldsymbol{\alpha}\) under the framework of multidimensional tensor sparse representation \cite{ref31} as follows
\begin{equation}
    \boldsymbol{\alpha} = \boldsymbol{\beta} \times_{1}\textbf{\textit{D}}_{1}\times_{2}\textbf{\textit{D}}_{2}\times_{3}\textbf{\textit{D}}_{3}\times_{4}\textbf{\textit{D}}_{4},\quad s.t.\ \left \|\boldsymbol{\beta}\right\|_{0}\le S
\end{equation}
where \(\boldsymbol{\beta}\in R^{I_{1}\times I_{2} \times I_{3} \times I_{4}}\) is a 4-order sparse tensor, \(\textbf{\textit{D}}_{1}\in R^{M\times I_{1}}\), \(\textbf{\textit{D}}_{2}\in R^{N\times I_{2}}\), \(\textbf{\textit{D}}_{3}\in R^{Q\times I_{3}}\) and \(\textbf{\textit{D}}_{4}\in R^{T\times I_{4}}\) are learned dictionary matrices, respectively. Here \(I_{1}\le M\), \(I_{2}\le N\), \(I_{3}\le Q\), \(I_{4}\le T\) and \textit{S} is the sparsity indicating the number of non-zero elements in \(\boldsymbol{\beta}\).

Combining (1), (2) and (3), the proposed kernel space-based multidimensional tensor sparse coding problem is formulated as
\begin{equation}
    \begin{split}
    \min_{\boldsymbol{\alpha},\boldsymbol{{\beta}}}  \frac{1}{2} &\; \left \|\tau^{-1}(\textbf{\textit{K}}\tau(\boldsymbol{\alpha})) -\textbf{\textit{y}}\right \|_{F}^{2} +\lambda_{1}( \frac{1}{2} ||\boldsymbol{\alpha} - \boldsymbol{\beta}\times_{1}\textbf{\textit{D}}_{1}    \\
    &\; \times_{2} \textbf{\textit{D}}_{2} \times_{3} \textbf{\textit{D}}_{3} \times_{4} \textbf{\textit{D}}_{4} ||_{F}^{2}+\lambda_{2}\left \| \boldsymbol{\beta}\right \|_{0})
    \end{split}
\end{equation}
% \begin{equation}
%     % \begin{split}
%     \min_{\boldsymbol{\alpha},\boldsymbol{{\beta}}}  \frac{1}{2} \left \|\tau^{-1}(\textbf{\textit{K}}\tau(\boldsymbol{\alpha})) -\textbf{\textit{y}}\right \|_{F}^{2} +\lambda_{1}( \frac{1}{2} ||\boldsymbol{\alpha} - \boldsymbol{\beta}\times_{1}\textbf{\textit{D}}_{1}
%     \times_{2} \textbf{\textit{D}}_{2} \times_{3} \textbf{\textit{D}}_{3} \times_{4} \textbf{\textit{D}}_{4} ||_{F}^{2}+\lambda_{2}\left \| \boldsymbol{\beta}\right \|_{0}) 
%     % \end{split}
% \end{equation}
where \(\tau^{-1}(\cdot )\) represents the operation of transforming a matrix of size \(MNQ\times T\) into a matrix of size \(M\times N\times Q\times T\), the first term and the second term stand for the data-fitting terms in the original space and the kernel space, respectively. \(\left \| \boldsymbol{\beta} \right \|_{0}\) stands for the sparsity constraint term, \(\lambda_{1}\) and \(\lambda_{2}\) are parameters to control the fidelity and sparsity, respectively.

\subsection{KMDS model optimization}
Referencing \cite{ref32}, (4) can be optimized using a greedy approach, successively learning \(\boldsymbol{\alpha}\) from \textit{\textbf{y}} and sparse tensor \(\boldsymbol{\beta}\) from \(\boldsymbol{\alpha}\). Let \(\textbf{\textit{K}}^{+}\) denote the pseudo-inverse matrix of \textbf{\textit{K}}, then \(\boldsymbol{\alpha}\) may be calculated by
\begin{equation}
    \boldsymbol{\alpha} = \tau^{-1} (\textbf{\textit{K}}^{+}\tau(\textbf{\textit{y}}))
\end{equation}

After obtaining \(\boldsymbol{\alpha}\), next we optimize \(\boldsymbol{\beta}\). First, we relax the non-convex \(l_{0}\) norm constraint in (4) to the \(l_{1}\) norm to obtain a convex optimization problem as follows
\begin{equation}
    \min_{\boldsymbol{\beta}} \frac{1}{2}\left \|\boldsymbol{\alpha} - \boldsymbol{\beta}\times_{1}\textbf{\textit{D}}_{1}\times_{2} \textbf{\textit{D}}_{2}\times_{3} \textbf{\textit{D}}_{3}\times_{4} \textbf{\textit{D}}_{4}\right \|_{F}^{2}+\lambda_{2}\left \| \boldsymbol{\beta}\right \|_{1}
\end{equation}
Then this problem can be optimized using Tensor-based Iterative Shrinkage Thresholding Algorithm (TISTA) \cite{ref31} by solving
% \begin{flalign}
%     &\ \boldsymbol{\beta}_{k} =  S_{\lambda_{2} /L_{k}}(\boldsymbol{\beta}_{k-1}-\frac{1}{L_{k}}(\boldsymbol{\beta}_{k-1}\times_{1}\textbf{\textit{D}}_{1}^{T}\textbf{\textit{D}}_{1}\times_{2}\textbf{\textit{D}}_{2}^{T}\textbf{\textit{D}}_{2}  \notag\\
%     &\ \qquad \times_{3}\textbf{\textit{D}}_{3}^{T}\textbf{\textit{D}}_{3} \times_{4}\textbf{\textit{D}}_{4}^{T}\textbf{\textit{D}}_{4} -\boldsymbol{\alpha} \times_{1}\textbf{\textit{D}}_{1}^{T}\times_{2}\textbf{\textit{D}}_{2}^{T}\times_{3}\textbf{\textit{D}}_{3}^{T}\times_{4}\textbf{\textit{D}}_{4}^{T})) &
% \end{flalign}
\begin{equation}
    \begin{split}
        \boldsymbol{\beta}_{k} &\; =  S_{\lambda_{2} /L_{k}}(\boldsymbol{\beta}_{k-1}-\frac{1}{L_{k}}(\boldsymbol{\beta}_{k-1}\times_{1}\textbf{\textit{D}}_{1}^{T}\textbf{\textit{D}}_{1}\times_{2}\textbf{\textit{D}}_{2}^{T}\textbf{\textit{D}}_{2}  \\
        &\; \times_{3}\textbf{\textit{D}}_{3}^{T}\textbf{\textit{D}}_{3} \times_{4}\textbf{\textit{D}}_{4}^{T}\textbf{\textit{D}}_{4} -\boldsymbol{\alpha} \times_{1}\textbf{\textit{D}}_{1}^{T}\times_{2}\textbf{\textit{D}}_{2}^{T}\times_{3}\textbf{\textit{D}}_{3}^{T}\times_{4}\textbf{\textit{D}}_{4}^{T}))
    \end{split}
\end{equation}
where \(S_{\lambda_{2} /L_{k}}(\cdot)=sign(\cdot)\max(\left |\cdot\right|- \lambda_{2} /L_{k}, 0)\) is the soft-thresholding operator,  $L_{k}>0$ is a Lipschitz constant and \(\textbf{\textit{D}}_{i}^{T}\) represents the transpose of \(\textbf{\textit{D}}_{i}\). To update $\boldsymbol{\beta}$, (7) can be rewritten as
\begin{equation}
    \begin{split}
        \boldsymbol{\beta}_{k}  =  &\; S_{\lambda_{2} /L_{k}}(\boldsymbol{\beta}_{k-1}+\frac{1}{L_{k}}(\boldsymbol{\alpha} -\boldsymbol{\beta}_{k-1}\times_{1}\textbf{\textit{D}}_{1} \times_{2}\textbf{\textit{D}}_{2} \\
        &\; \times_{3}\textbf{\textit{D}}_{3}\times_{4}\textbf{\textit{D}}_{4})\times_{1}\textbf{\textit{D}}_{1}^{T}\times_{2}\textbf{\textit{D}}_{2}^{T}\times_{3}\textbf{\textit{D}}_{3}^{T}\times_{4}\textbf{\textit{D}}_{4}^{T})
    \end{split}
\end{equation}
The optimization of \(\boldsymbol{\beta}\) in (8) can be decoupled into three steps and sequentially connected to form a trainable network. Then, (8) becomes
\begin{align}
    \boldsymbol{\varsigma}_{k}= &\; \boldsymbol{\alpha} - \boldsymbol{\beta}_{k-1}\times_{1}\textbf{\textit{D}}_{1}\times_{2}\textbf{\textit{D}}_{2}\times_{3}\textbf{\textit{D}}_{3}\times_{4}\textbf{\textit{D}}_{4} \\
    \boldsymbol{\xi}_{k}= &\; \boldsymbol{\beta}_{k-1}+\frac{1}{L_{k}} \boldsymbol{\varsigma}_{k}\times_{1}\textbf{\textit{D}}_{1}^{T}\times_{2}\textbf{\textit{D}}_{2}^{T}\times_{3}\textbf{\textit{D}}_{3}^{T}\times_{4}\textbf{\textit{D}}_{4}^{T} \\
    \boldsymbol{\beta}_{k}= &\; S_{\lambda_{2} /L_{k}}(\boldsymbol{\xi}_{k})
\end{align}
The initialization of \(\boldsymbol{{\beta}}\) can be set to \(\boldsymbol{{\beta}}_{0}=\boldsymbol{{\alpha}}\times_{1} \textbf{\textit{D}}^{T}_{1} \times_{2} \textit{\textbf{D}}^{T}_{2} \times_{3} \textbf{\textit{D}}^{T}_{3}\times_{4} \textbf{\textit{D}}^{T}_{4}\). 

After computing \(\boldsymbol{\beta}_{k}\), \(\hat{\boldsymbol{\alpha}}\) can be obtained using the following formula
\begin{equation}
    \hat{\boldsymbol{\alpha}} = {\boldsymbol{\beta}}_{k}\times_{1}\textbf{\textit{D}}_{1}\times_{2}\textbf{\textit{D}}_{2}\times_{3}\textbf{\textit{D}}_{3}\times_{4}\textbf{\textit{D}}_{4}
\end{equation}

 Finally, the denoised PET image \(\hat{\textbf{\textit{x}}}\) can be obtained using (2). 
 
 % Algorithm 1 provides the process of the KMDS method. 
 The optimization of \(\boldsymbol{\alpha}\) and \(\boldsymbol{\beta}\) mentioned above requires computationally inefficient iterations and complex fine-tuning of hyperparameters, such as \(\lambda_{1}\), \(\lambda_{2}\) and \(L_{k}\), which limit the generalizability of the KMDS algorithm. An alternative method is to unfold the iterative optimization steps of \(\boldsymbol{\beta}\) as network layers to automatically learn the sparse dictionaries and hyperparameters \cite{ref36}. However, there are noticeable differences in image quality and quantitation across frames of dynamic PET. This makes simple dictionary matrices with limited representation ability unable to address the challenging issue, resulting in obvious noise and bias in the shorter scan denoised using \cite{ref36} directly. In addition, the kernel matrix \textbf{\textit{K}} of an image is usually too large to calculate its pseudo-inverse \(\textbf{\textit{K}}^{+}\). To address these issues, we propose a neural KMDS-Net, as described in the next section.

\subsection{Neural KMDS-Net}
The kernel representation \(\tau^{-1} (\textbf{\textit{K}}^{+}\tau(\textbf{\textit{y}}))\) of \textbf{\textit{y}} in (5) may be generalized to multiple image convolutions using learned local filters, i.e.
\begin{equation}
    \boldsymbol{\alpha}=\textbf{\textit{y}}\circ\textbf{\textit{c}}_{1}^{n}\circ\textbf{\textit{c}}_{2}^{n}\circ\cdots \circ\textbf{\textit{c}}_{m}^{n}
\end{equation}
where $\textbf{\textit{y}} \circ \textbf{\textit{c}}_{i}^{n}=\textbf{\textit{c}}_{i}^{n}(\textbf{\textit{y}})$, \(\textbf{\textit{c}}_{m}^{n}\) denotes the \textit{m}-th convolutional layer with output filter number $n$. To prevent gradient explosion, batch normalization (BN) \cite{ref33} layer is added after each convolution layer. Let \(\textbf{\textit{b}}_{m}\) represent the \textit{m}-th BN layer, we have
\begin{equation}
    \boldsymbol{\alpha}=\boldsymbol{\Theta}(\textbf{\textit{y}})
\end{equation}
where $\boldsymbol{\Theta}(\textbf{\textit{y}})=\textbf{\textit{y}}\circ\textbf{\textit{c}}_{1}^{n}\circ\textit{\textbf{b}}_{1}\circ\textbf{\textit{c}}_{2}^{n}\circ\textit{\textbf{b}}_{2}\circ\cdots\circ\textbf{\textit{c}}_{m}^{n}\circ\textbf{\textit{b}}_{m}$.

Considering the strong representation ability of DL, we use convolutional neural networks to substitute dictionary matrices to improve the performance of KMDS method by learning the sparse representation $\boldsymbol{\beta}$ from \(\boldsymbol{\alpha}\) more accurately. Then, (9)-(11) can be written as follows
\begin{align}
     \boldsymbol{\varsigma}_{k}= &\; \boldsymbol{\alpha} - \boldsymbol{\Phi}_{k}(\boldsymbol{\beta}_{k-1}) \\
    \boldsymbol{\xi}_{k}= &\; \boldsymbol{\beta}_{k-1}+ \boldsymbol{\Gamma}_{k}(\boldsymbol{\varsigma}_{k}) \\
    \boldsymbol{\beta}_{k}= &\; S_{\boldsymbol{\rho_{k}}}(\boldsymbol{\xi}_{k})
\end{align}
where \(\boldsymbol{\Phi}_{k}\) and \(\boldsymbol{\Gamma}_{k}\) both represent a CNN with four cascaded convolutional layers. A BN layer is added after each convolutional layer. Each convolutional layer corresponds to a dictionary matrix. \(1/L_{k}\) is implicitly merged to \(\boldsymbol{\Gamma }_{k}\). \(\boldsymbol{\rho}_{k}\in R^{1\times 1\times 1\times I_{4}}\) denotes the soft thresholding parameter tensor for the \textit{k}-th iteration optimization of the parameter \(\boldsymbol{\beta}\). The soft-thresholding operator can be written as \(S_{\boldsymbol{\rho}_{k}}(\boldsymbol{\xi}_{k})=\max(\boldsymbol{\xi}_{k}-\boldsymbol{\rho}_{k}, 0)-\max(-\boldsymbol{\xi}_{k}-\boldsymbol{\rho}_{k}, 0)\). \(\max(\cdot)\) can be replaced by a rectified linear units (ReLU) layer, and we have \(S_{\boldsymbol{\rho}_{k}}(\boldsymbol{\xi}_{k})={\rm ReLU}(\boldsymbol{\xi}_{k}-\boldsymbol{\rho}_{k})-{\rm ReLU}(-\boldsymbol{\xi}_{k}-\boldsymbol{\rho}_{k})\). 
Furthermore, the dictionary matrices used in the initialization of $\boldsymbol{\beta}_{0}$ may also be substituted with convolutional layers. We have $\boldsymbol{\beta}_{0}=\boldsymbol{\Lambda}(\boldsymbol{\alpha})$, where $\boldsymbol{\Lambda}$ and $\boldsymbol{\Phi}_{k}$ share the same network architecture. Therefore, the optimization of the sparse representation \(\boldsymbol{\beta}\) may be implemented by a fully convolution neural network which may achieve better results by implicitly embedding explicit parameters into the neural network training process. We compose (15), (16) and (17) into a multidimensional sparse block (MDSB) and stack it for $k$ times to form a sub-network for multidimensional sparse coding.

After obtaining \(\boldsymbol{\beta}_{k}\), \(\hat{\boldsymbol{\alpha}}\) in (12) may also be represented by a neural network as follows
\begin{equation}
    \hat{\boldsymbol{\alpha}} = \boldsymbol{\Psi}({\boldsymbol{\beta}}_{k})
\end{equation}
where \(\boldsymbol{\Psi}\) , similar to \(\boldsymbol{\Phi }_{k}\), is also a four-layer CNN. Finally, the kernel matrix $\textbf{\textit{K}}$ in (2) may also be replaced by multiple convolutional layers. The denoised \(\hat{\textbf{\textit{x}}}\) may be obtained by
\begin{equation}
    \hat{\textbf{\textit{x}}} = \textbf{\textit{O}}(\hat{\boldsymbol{\alpha}})
\end{equation}
where \(\textbf{\textit{O}}(\hat{\boldsymbol{\alpha}})=\hat{\boldsymbol{\alpha}}\circ\textbf{\textit{d}}_{1}^{n}\circ\textit{\textbf{b}}_{1}\circ\textbf{\textit{d}}_{2}^{n}\circ\textit{\textbf{b}}_{2}\circ\cdots\circ\textbf{\textit{d}}_{m}^{n}\circ\textbf{\textit{b}}_{m}\) and \(\textbf{\textit{d}}_{m}^{n}\) denotes the $m$-th convolutional layer with output filter number $n$. We refer to this model as the neural KMDS-Net. The network architecture is shown in Fig. \ref{fig:1}.

\subsection{Network training loss}
The mean squared error (MSE) between the output \(\hat{\textbf{\textit{x}}}\) of neural KMDS-Net and the reference image \textbf{\textit{x}} is used to form the training loss function
\begin{equation}
    L=\left \| \hat{\textbf{\textit{x}}}-\textbf{\textit{x}} \right \|_{2}^{2}
\end{equation}

\subsection{Evaluation metrics}
Peak signal-to-noise ratio (PSNR) and the structural similarity (SSIM) are used to compare different image denoising methods. The definition of the image PSNR is as follows:
\begin{equation}
    \rm{PSNR} = 10 \rm{log}_{10}(\frac{(max(\textit{c}^{true}))^{2}}{\left \| \hat{\textit{c}} - \textit{c}^{true} \right \|^{2}}  )
\end{equation}
where \(c^{\rm true}\) and \(\hat{c}\) are the ground truth and denoised images, respectively. \(\rm{max}(\textit{c}^{true})\) represents the maximum pixel value of \(c^{\rm true}\). 

The SSIM is defined as follows:
\begin{equation}
    \rm{SSIM} = \frac{2\mu_{\hat{c}}\mu_{c^{\rm true}}+C_{1}}{\mu_{c^{\rm true}}^{2}+\mu_{\hat{c}}^{2}+C_{1}}\times \frac{2\sigma_{\hat{c},c^{\rm true}}+C_{2}}{\sigma_{\hat{c}} \sigma_{c^{\rm true}}+C_{2}}
\end{equation}
where \(\mu\) denotes the mean of the image and \(\sigma\) represents the standard deviation of the image. Specifically, \(\rm{C}_{1} = 0.01\times max(c^{\rm true})\) and \(\rm{C}_{2}=0.03\times max(c^{\rm true})\).

The ensemble bias of the mean intensity and background normalized standard deviation (NSD) in regions of interest (ROIs) are also computed to evaluate ROI quantification and measure the noise, 
\begin{align}  
    {\rm Bias} = &\; \frac{|\bar{c}-c^{\rm true}|}{c^{\rm true}} \\
    {\rm NSD}=&\;\frac{1}{N_{r}} {\textstyle \sum_{i=1}^{N_{r}}}\frac{1}{\bar{b_{i}}}\sqrt{\frac{1}{N_{b}-1}{\textstyle \sum_{j=1}^{N_{b}}(b_{i,j}-\bar{b_{i}})^{2}}}   
\end{align}
where \(\bar{c}=\frac{1}{N_{r} }  \sum_{i=1}^{N_{r}}\bar{c_{i}}\) denotes the mean of \(N_{r}\) realizations. \(\bar{c_{i}}\) is the mean ROI uptake in the \textit{i}-th realization and \(N_{r}=10\) in this study. \(b_{i,j}\) is the \textit{j}-th voxel intensity of the background ROI in the $i$-th realization. \(\bar{b_{i}}\) is the mean voxel intensity and \(N_{b}\) denotes the number of voxels in the background ROI.

\subsection{Implement details}
Neural KMDS-Net is implemented in Pytorch and trained on a NVIDIA Tesla V100 GPU. 
Training images at different frames are divided by the global maximum voxel value across all frames. The Adam algorithm is used with a initial learning rate \(5\times 10^{-3}\), which is multiplied by 0.35 for every 80 epochs. The batch size is 2 and the model is trained for 300 epochs until convergence. 3D convolutional layers and 3D BN are used in neural KMDS-Net. Each convolutional layer in neural KMDS-Net is with kernel size 3, stride 1 and padding 1. The network's input and output are tensors of size $B \times T \times Q \times M \times N$, where $B$ represents the batch size. $m$ and $n$ in (13) and (19) are set to 2 and $T$, respectively. The number of input and output channels for each convolutional layer in $\boldsymbol{\Lambda}$, \(\boldsymbol{\Phi}_{k}\), \(\boldsymbol{\Gamma}_{k}\) and $\boldsymbol{\Psi}$ is set to \textit{T}. The number of unfolding $k$ is set to 20. The model parameters of neural KMDS-Net in simulation and real studies are 0.7 M and 0.9 M respectively. The same network architecture and most training hyperparameters are used for both the simulated and clinical datasets. The only exception is the number of input and output channels in the convolutional layers, which is determined by the frame number of the input dynamic PET data.

\subsection{Reference methods}
We compare the proposed neural KMDS-Net with the standard ML-EM method, the kernelized expectation maximization (KEM) method \cite{ref5}, the ResNet direct denoising method \cite{ref34}, the Pix2Pix-based generative adversarial network (GAN) direct denoising method \cite{ref35}, the subspace-based multidimensional sparse (SMDS)-Net iterative unrolling network method \cite{ref36} and the DDPM-PET method \cite{ref37}.
In the KEM method, three 20-minute composite frames are used as the prior images. \(\sigma\) is set to 1 in the radial Gaussian kernel function  and the number of nearest neighbors is set to 50 in $k$NN to generate the sparse kernel matrix \textbf{\textit{K}}.
For the ResNet direct denoising method, the number of output filters in the first convolutional layer is set to 32 and the number of residual block is set to 9.
In the Pix2Pix-based GAN, the generator uses the same ResNet structure mentioned above and the number of output channels for the first convolutional layer of the discriminator is 64. The input noisy image is added to the discriminator as a conditional term. The weight for the discriminator loss is empirically set to 0.03 based on our previous work \cite{ref38} and experimental analysis.
3D convolution layers and 3D BN are used in ResNet and GAN.
The DDPM-PET method only uses the PET images as the input for the network and is implemented using the modified open-source PALETTE framework \cite{ref39}. The batch size is set to 16. The base channel count is set to 32 and the number of residual blocks is set to 1. Other model parameters are configured to be consistent with the parameters for the colorization task. 
The SMDS-Net method is implemented through the open-source code provided by the authors. Some parameters, such as patch size (\(M\times N\)), cube size ([5, 5, 5]), number of dictionaries ([9, 9, 9]) and unfolding numbers (3), are reconfigured to achieve better results in dynamic PET image denoising.

\section{Simulation validation}
\label{sec:sim}
\subsection{Simulation setup}
\begin{figure*}[t]%[th!b]
    \centering
        \includegraphics[width=1\textwidth]{"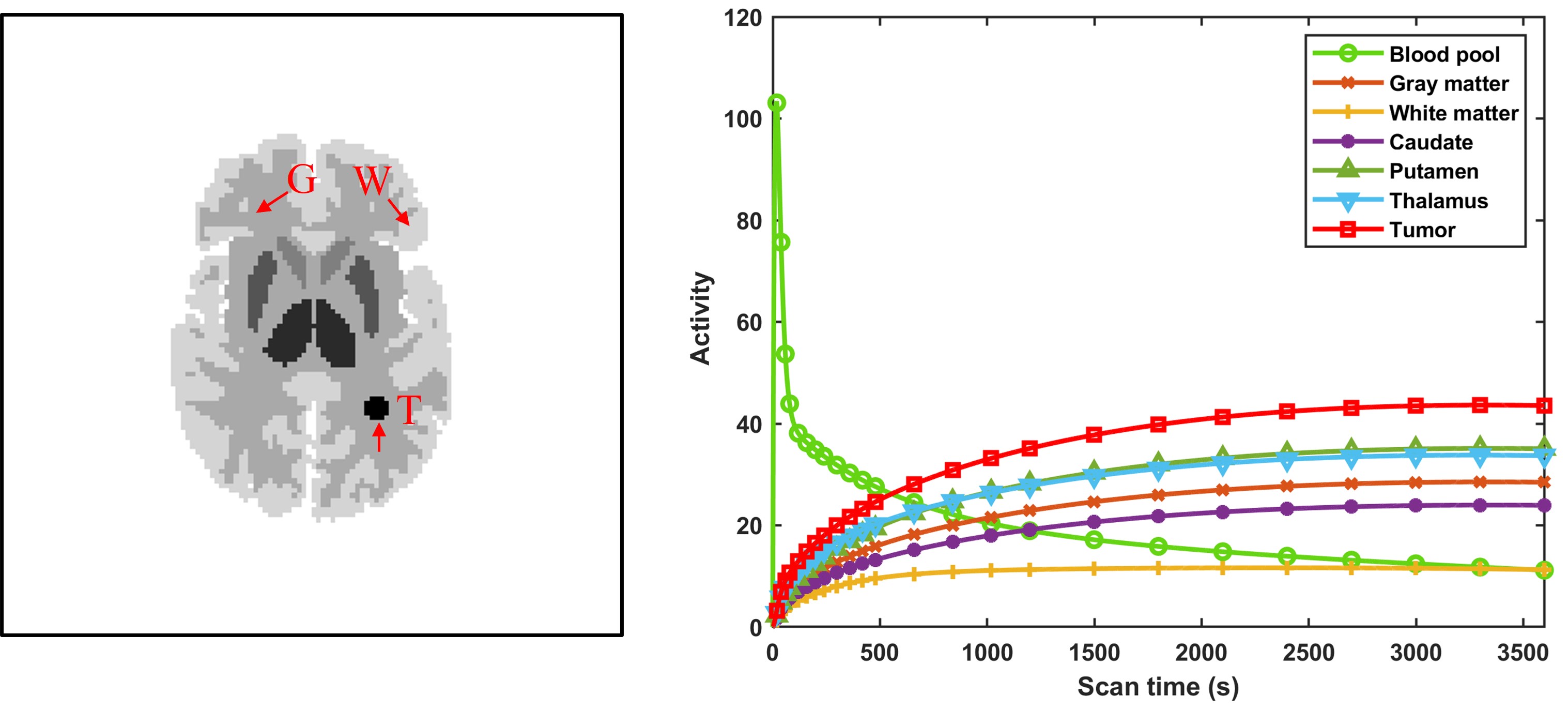"}
        \caption{The Zubal brain phantom (left) and examples of time activity curves (right) used in the simulation studies. ‘G’ represents the gray matter ROI, 'W' represents the white matter ROI and ‘T’ is the tumor ROI.}
    \label{fig:2}
\end{figure*}

Dynamic $^{18}$F-FDG scans are simulated for a PET/CT scanner using a \(128 \times 128 \times 8\) Zubal head phantom as shown in Fig. \ref{fig:2}. The phantom is composed of gray matter, white matter, caudate, putamen, thalamus and a tumor. The sampling protocol is set to 24 time frames: \(4 \times 20\)s, \(4 \times 40\)s, \(4 \times 60\)s, \(4 \times 180\)s, and \(8 \times 300\)s.

The time activity curves (TAC) of different regions are generated using a three compartmental model with a blood input function. For data augmentation, each kinetic parameter follows a Gaussian variable with coefficient of variation equal to 20\% of the mean values. The mean values of kinetic parameters are shown in Table \ref{tab1}. The TACs using the mean kinetic parameters are shown in Fig. \ref{fig:2}.
The total number of events over 1 hours is 1.19 billion.
Translation and scaling are applied to the phantom for data augmentation. The ground-truth images are first forward-projected to generate noise-free sinograms. Poisson noise is then introduced. The random and scatter events are
simulated using Monte Carlo simulation and account for approximately 30\% statistics of the noiseless data. For the training data, we generate 2000 noisy-clean PET image pairs, with each image being of size \(128 \times 128 \times 8 \times 24\). For the test data, in total 600 ($60\times 10$) pairs are generated, with 10 noisy realizations simulated for each true image.

\begin{table}[t]%\scriptsize%[th!b]
    \caption{The mean values of the simulated kinetics parameters for different organs. \textit{V} stands for blood volume ratio.} \label{tab1}
    \centering
    \begin{tabular}{ l c c c c c c}
        \hline\hline
         & \(\textit{K}_{1}\) & \(\textit{k}_{2}\) & \(\textit{k}_{3}\) & \(\textit{k}_{4}\) & \textit{V} \\
        \hline
        Gray matter & 0.080 & 0.140 & 0.170 & 0.013 & 0.103 \\
        White matter & 0.050 & 0.110 & 0.050 & 0.006 & 0.026 \\
        Caudate  & 0.070 & 0.170 & 0.190 & 0.016 & 0.101 \\
        Putamen & 0.090 & 0.160 & 0.170 & 0.010 & 0.092\\
        Thalamus & 0.110 & 0.160 & 0.140 & 0.012 & 0.152\\
        Tumor & 0.130 & 0.100 & 0.150 & 0.015 & 0.173\\
        \hline\hline
    \end{tabular}
\end{table}

\begin{figure*}[t]%[th!b]
    \centering
        \includegraphics[width=1.0\textwidth]{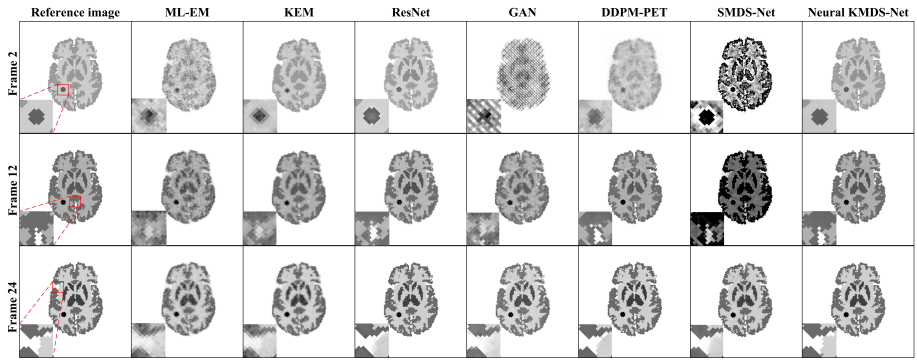}
        \caption{True activity images and denoised images by different denoising methods in the simulation study. The window size of frame 2 (top row), frame 12 (middle row) and frame 24 (bottom row) is set to [0, 0.01], [0, 0.1] and [0, 1.0], respectively.}
    \label{fig:3}
\end{figure*}

\begin{figure*}[t]%[th!b]
    \centering
        \includegraphics[width=1\textwidth]{"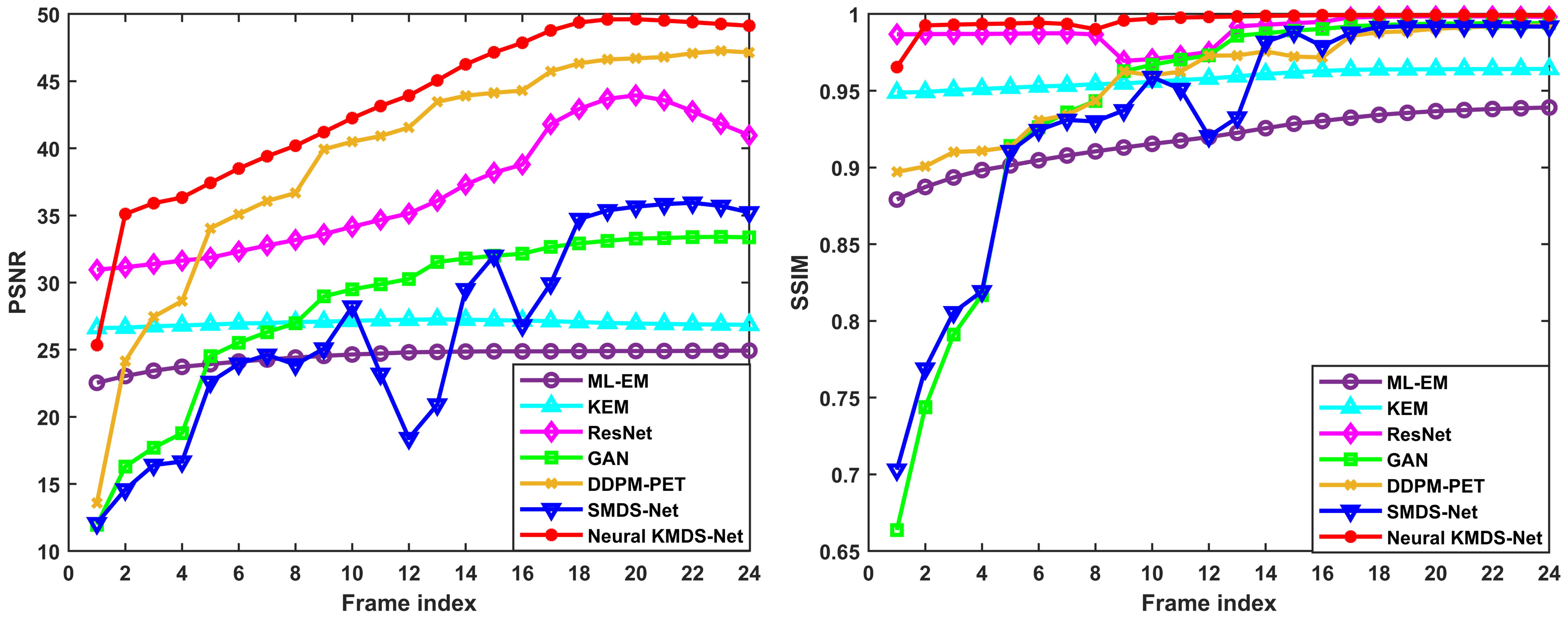"}
        \caption{Plots of image PSNR and SSIM of all time frames for different denoising methods.}
    \label{fig:4}
\end{figure*}

\begin{figure*}[t]%[th!b]
    \centering
        \includegraphics[width=1\textwidth]{"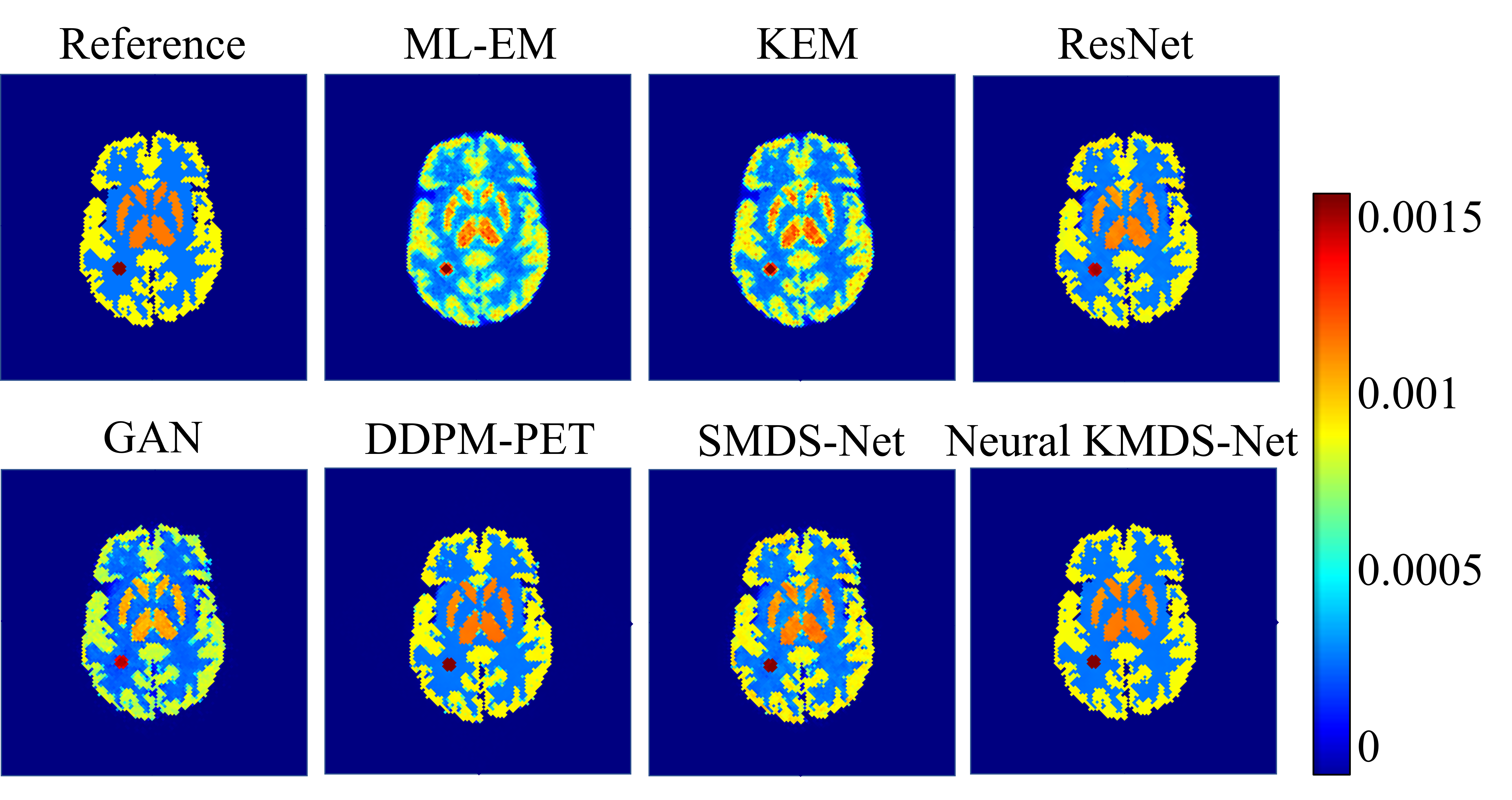"}
        \caption{Parametric images of $K_{i}$ generated from the images in Fig. \ref{fig:4}, estimated using Reference, ML-EM, KEM, ResNet, GAN, DDPM-PET, SMDS-Net and neural KMDS-Net.
}
    \label{fig:zubal-k}
\end{figure*}

\begin{table*}[t]\scriptsize%[th!b]
    \caption{Quantitative comparison of different organs by different methods in frame 2 and 24 from the simulation study. Mean PSNR is reported. Best results are marked in \textbf{BOLD}.} \label{tab2}
    \centering
    \setlength{\tabcolsep}{1.9pt}
    \begin{tabular}{ l c c c c c c c c c c c c}
        \hline\hline
        \multirow{2}{*}{Local PSNR} & \multicolumn{2}{c}{Gray matter} & \multicolumn{2}{c}{White matter} & \multicolumn{2}{c}{Caudate} & \multicolumn{2}{c}{Putamen} & \multicolumn{2}{c}{Thalamus} & \multicolumn{2}{c}{Tumor}  \\
        \cline{2-13}
        & 2 & 24 & 2 & 24 & 2 & 24 & 2 & 24 & 2 & 24 & 2 & 24 \\
        \hline
        ML-EM & 31.919 & 32.675 & 26.216 & 28.486 & 28.622 & 31.083 & 38.714 & 40.125 & 37.191 & 38.453 & 34.714 & 38.110 \\
        KEM & 32.564 & 34.395 & 31.589 & 30.892 & 33.806 & 32.632 & 40.637 & 41.348 & 41.768 & 40.571 & 37.232 & 39.741 \\
        ResNet & 42.473 & 55.042 & 35.498 & 46.339 & 39.105 & 47.854 & 49.134 & 57.286 & 44.261 & 53.892 & 42.169 & 52.777 \\
        GAN & 28.598 & 52.911 & 19.585 & 35.603 & 19.777 & 42.125 & 34.369 & 47.669 & 31.743 & 45.195 & 29.121 & 43.698 \\
        DDPM-PET & 31.780 & 56.929 & 28.303 & 50.809 & 30.110 & 53.891 & 41.318 & 60.734 & 39.622 & 59.685 & 37.672 &  60.957 \\
        SMDS-Net & 26.981 & 45.924 & 7.401 & 38.833 & 14.423 & 40.547 & 20.575 & 48.806 & 17.481 & 49.987 & 16.004 & 47.962 \\
        Neural KMDS-Net & \textbf{55.654} & \textbf{63.776} & \textbf{39.864} & \textbf{52.535} & \textbf{42.014} & \textbf{54.660} & \textbf{49.467} & \textbf{64.130} & \textbf{50.223} & \textbf{63.181} & \textbf{45.412} & \textbf{63.035} \\
        \hline\hline
    \end{tabular}
\end{table*}

\begin{table*}[t]\scriptsize%[th!b]
    \caption{Quantitative comparison of model complexity (parameters, FLOPs) and inference efficiency (time) for the evaluated models using the Zubal datasets. Best results are marked in \textbf{BOLD}.} \label{tab-para}
    \centering
    \setlength{\tabcolsep}{1.9pt}
    \renewcommand{\arraystretch}{1.5}
    \begin{tabular}{ l c c c c c c c c c c c c}
        \hline\hline
        Complexity \& Efficiency & ML-EM & KEM & ResNet & GAN & DDPM-PET & SMDS-Net & Neural KMDS-Net  \\
        \hline
        Model Parameters & - & - & 10.5 M & 10.5 M & 14.0 M & \textbf{0.002 M} & 0.7 M \\
        Computational Complexity & - & - & \textbf{0.5 GFLOPs} & \textbf{0.5 GFLOPs} & 77.7 GFLOPs & - & 11 GFLOPs \\
        Inference time & - & - & 0.02 s & 0.02 s & 72 s & 0.39 s & \textbf{0.01 s} \\
        \hline\hline
    \end{tabular}
\end{table*}

\subsection{Comparison of Image Quality}
Fig. \ref{fig:3} shows the ground-truth images and the denoised images by seven different methods for frame 2, frame 12 and frame 24 in the simulation. It may be observed that ML-EM has the highest noise among all. The regular KEM achieves better results compared to ML-EM. ResNet suppresses noised well and preserves fine details. GAN yieldes very poor results for frame 2 due to the significant differences in pixel intensity distribution between frames, which exacerbated the difficulty of the model's learning process. For frame 2, DDPM-PET is noisier than ResNet. SMDS-Net yieldes significantly biased results for frame 2 and frame 12 partially due to the limited learning capacity of the dictionary matrix. In comparison, benefiting from the interpretable neural network, the proposed neural KMDS-Net achieves the best image quality and demonstrated best detail preservation.

Fig. \ref{fig:4} further shows quantitative assessment for all time frames denoised by different methods. We may observe that the neural KMDS-Net is with a higher PSNR for frame 1 and outperforms all compared baselines in terms of PSNR and SSIM for other frames. The quantitative assessment of six organs using different methods for frame 2 and frame 24 is summarized in Table \ref{tab2}. The neural KMDS-Net shows a significant improvement in PSNR over other methods in all organ ROIs. 

Table \ref{tab-para} presents a quantitative comparison of model complexity and inference efficiency for different DL methods on the Zubal dataset. ResNet (10.5 M) and GAN (10.5 M) exhibit the largest number of parameters and the lowest computational complexity. DDPM-PET (14.0 M) requires the longest inference time. SMDS-Net (0.002 M) has the fewest model parameters but relatively high inference time. In contrast, Neural KMDS-Net (0.7 M) achieves a favorable trade-off between model parameters and computational complexity, while delivering the fastest inference speed.

To evaluate the temporal fidelity for dynamic PET, we present parametric maps of $K_{i}$ in Fig. \ref{fig:zubal-k}. ML-EM shows severe noise, while KEM exhibits limited noise suppression. GAN and SMDS-Net reduce noise but generate obvious artifacts. ResNet removes most noise yet introduces bias in tumor regions. DDPM-PET achieves strong noise reduction but contains scattered pixels with large intensity bias. In contrast, the parametric image generated by neural KMDS-Net exhibits more complete and regular tissue structures with improved structural fidelity.

\begin{figure*}[t]%[th!b]
    \centering
        \includegraphics[width=1.0\textwidth]{"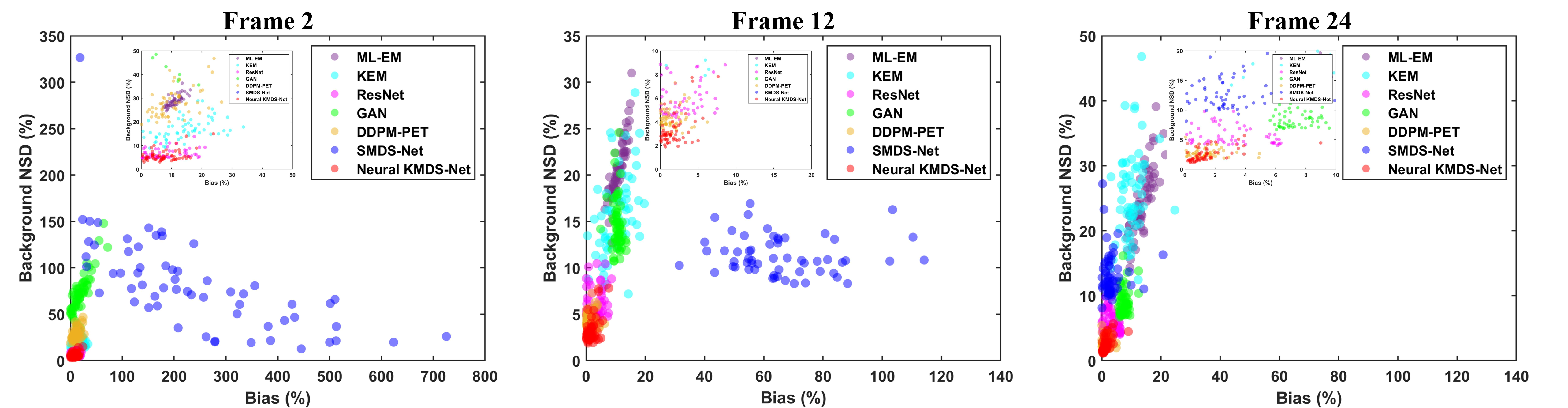"}
        \caption{Plots of bias-NSD for tumor ROI quantification in frame 2, 12 and 24. White matter is selected as the background to compute NSD. A zoom-in is included for each frame.}
    \label{fig:5}
\end{figure*}

\begin{figure*}[t]%[th!b]
    \centering
        \includegraphics[width=1.0\textwidth]{"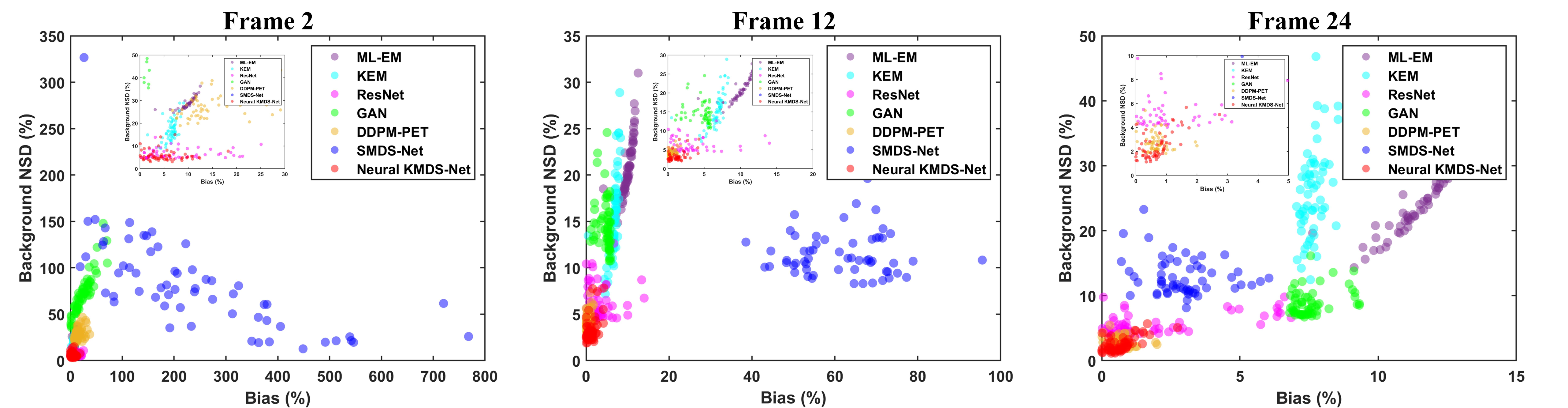"}
        \caption{Plots of bias-NSD for gray matter ROI quantification in frame 2, 12 and 24. White matter is selected as the background to compute NSD. A zoom-in is included for each frame.}
    \label{fig:6}
\end{figure*}

\subsection{Comparison of ROI quantification}
Figs. \ref{fig:5}-\ref{fig:6} show the trade-off between the absolute bias and NSD of different methods for ROI quantification in the tumor (Fig. \ref{fig:5}) and gray matter (Fig. \ref{fig:6}) regions. The two ROIs are the same as the anatomical
regions marked in Fig. \ref{fig:2}. Each point is obtained by calculating the average of the denoising results from 10 noisy realizations of a test image. For both the tumor ROI and gray matter quantification, SMDS-Net is with the highest bias in frame 2 and frame 12 but performs better than the GAN, KEM and ML-EM in frame 24. DDPM-PET demonstrates a remarkable superiority in frame 12 and frame 24 compared to other regular DL methods and traditional approaches. However, it exhibits relatively higher background noise in frame 2 compared to the KEM, ResNet and the proposed method. In comparison, the proposed neural KMDS-Net achieves the best bias-NSD trade-off among the other reference methods.

\begin{table}[t]%\scriptsize%[th!b]
    \caption{The PSNR of different depth $m$ of neural networks in kernel representation and image estimation modules on the Zubal head dataset. Best results are marked in \textbf{BOLD}.} \label{tab3}
    \centering
    \begin{tabular}{ c c c c c c c c}
        \hline\hline
        $m$ & 0 & 2 & 4 & 6 & 8 & 10 & 12 \\
        \hline
        Frame 2  & 32.43 & \textbf{35.05} & 33.68 & 33.36 & 31.31 & 30.82 & 33.23 \\
        Frame 12 & 40.74 & \textbf{42.11} & 40.48 & 39.67 & 39.64 & 39.67 & 38.93 \\
        Frame 24 & 44.05 & 45.22 & 45.06 & 44.63 & \textbf{45.40} & 44.54 & 44.64 \\
        \hline\hline
    \end{tabular}
\end{table}

\begin{table}[t]%\scriptsize%[th!b]
    \caption{The PSNR of different unfolding iteration $k$ on the Zubal head dataset. Best results are marked in \textbf{BOLD}.} \label{tab4}
    \centering
    \begin{tabular}{ c c c c c c c}
        \hline\hline
        $k$ & 5 & 10 & 15 & 20 & 25 & 30 \\
        \hline
        Frame 2  & 34.10 & 34.72 & 32.87 & \textbf{35.05} & 33.63 & 32.34 \\
        Frame 12 & 40.31 & 41.07 & 41.20 & \textbf{42.11} & 39.65 & 42.04 \\
        Frame 24 & 42.18 & 43.82 & 44.26 & 45.22 & 45.69 & \textbf{46.35} \\
        \hline\hline
    \end{tabular}
\end{table}

% \begin{table*}[t]%[th!b]
%     \caption{The PSNR of different sizes of \(\beta\) on the simulated dataset. Best results are marked in \textbf{BOLD}.} \label{tab5}
%     \centering
%     \begin{tabular}{ c l c c c c c c}
%         \hline\hline
%         $I_{1}, I_{2}, I_{3}, I_{4}$ & & 12 & 16 & 20 & 24 & 28 & 32 \\
%         \hline
%         & Frame 2  &  32.54 & 33.32 & 31.50 & \textbf{33.62} & 32.54 & 31.41 \\
%        $64 \times 64 \times 4$ & Frame 12 &  37.74 & 37.84 & 38.29 & 39.39 & \textbf{40.11} & 38.28 \\
%         & Frame 24 &  41.65 & 43.28 & 43.29 & \textbf{44.83} & 44.09 & 44.14 \\
%         \hline
%         & Frame 2 & 34.05 & 33.58 & 33.34 & \textbf{35.05} & 32.45 & 32.06 \\
%        $128 \times 128 \times 8$ & Frame 12 & 40.69 & 41.58 & 41.37 & \textbf{42.11} & 41.00 & 41.93 \\
%         & Frame 24 & 44.31 & 45.242 & 44.66 & 45.22 & 44.05 & \textbf{46.41} \\
%         \hline\hline
%     \end{tabular}
% \end{table*}

\begin{table}[t]%\scriptsize%[th!b]
    \caption{The PSNR of different sizes of \(\beta\) on the Zubal head dataset. Best results are marked in \textbf{BOLD}.} \label{tab5}
    \centering
    \begin{tabular}{ c c c c c}
        \hline\hline
        $I_{1}, I_{2}, I_{3}$ & $I_{4}$ & Frame 2 & Frame 12 & Frame 24 \\
        \hline
        \multirow{6}{*}{$64 \times 64 \times 4$} & 12 & 32.54 & 37.74 & 41.65 \\
        & 16 & 33.32 & 37.84 & 43.28\\
        & 20 & 31.50 & 38.29 & 43.29 \\
        & 24 & \textbf{33.62} & 39.39 & \textbf{44.83} \\
        & 28 & 32.54 & \textbf{40.11} & 44.09 \\
        & 32 & 31.41 & 38.28 & 44.14 \\
        \hline
        \multirow{6}{*}{$128 \times 128 \times 8$} & 12 & 34.05 & 40.69 & 44.31 \\
        & 16 & 33.58 & 41.58 & 45.24 \\
        & 20 & 33.34 & 41.37 & 44.66 \\
        & 24 & \textbf{35.05} & \textbf{42.11} & 45.22 \\
        & 28 & 32.45 & 41.00 & 44.05 \\
        & 32 & 32.06 & 41.93 & \textbf{46.41} \\
        \hline\hline
    \end{tabular}
\end{table}

\subsection{Effect of parameter \textit{m}}
We change the depth \textit{m} of neural networks in (13) and (19) from 0 to 12 with an interval of 2. The number of 0 indicates that neural KMDS-Net does not include kernel representation and image estimation modules. As shown in Table \ref{tab3}, increasing the depth of neural networks does not always allow for better denoising performance. This may be attributed to the straightforward concatenation architecture of neural networks. Therefore, the parameter $m$ is set to 2 considering the denoising performance.

\subsection{Effect of unfolding iteration $\textit{k}$}
$k$ denotes the number of iterations for the parameters updates in (15) to (17) and also controls the depth of the neural network modules of multidimensional sparse representation. Table \ref{tab4} shows the influence of the unfolding iteration $k$. A larger $k$ enforces the network to take advantage of the information between current iteration and historical iterations to optimize the parameters, implying that larger number should be chosen. However, increasing $k$ from 20 to 30, the denoising performance does not show a significant improvement. Considering the tradeoff between the inference time and denoising performance, $k$ is set as 20.

\subsection{Effect of different sizes of $\beta$ }
Here, we analyze the influence of the size of sparse tensor $\boldsymbol{\beta}$ in multidimensional sparse coding on dynamic PET denoising. For simplicity, we set $I_{1}=I_{2}$ and choose the value from the set of $\left \{64, 128\right \} $. $I_{3}$ is chosen from the set of $\left \{4, 8\right \} $. $I_{4}$ is changed from 12 to 32 with an interval of 4. When $I_{1}=I_{2}=64$ and $I_{3}=4$, the stride of the first convolutional layer in $\boldsymbol{\Lambda}$ and $\boldsymbol{\Gamma}$ is set to 2 and the first layer in $\boldsymbol{\Phi}$ and $\boldsymbol{\Psi}$ is replaced with a transposed convolutional layer. As shown in Table \ref{tab5}, using larger $I_{1}$, $I_{2}$, $I_{3}$ and $I_{4}$ enhance the representation ability of neural KMDS-Net and achieve better denoising performance. However, increasing $I_{4}$ from 24 to 32, the denoising performance changes are not significant. This is also consistent with the constraints on the size of $\boldsymbol{\beta}$ in (3).

\begin{figure*}[t]%[th!b]
    \centering
        \includegraphics[width=1.0\textwidth]{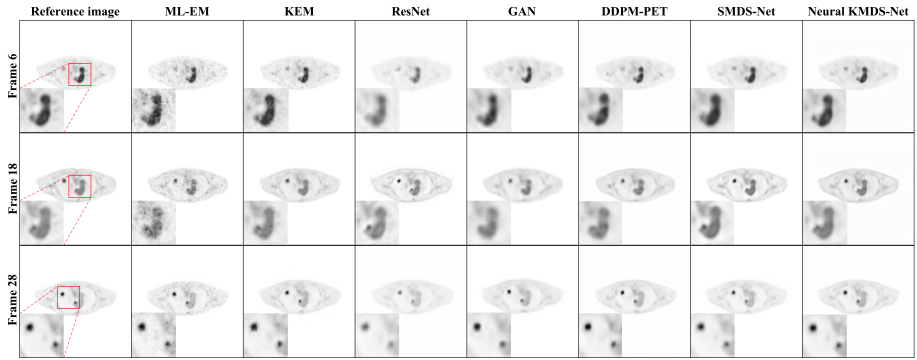}
        \caption{True activity images and denoised images by different methods in the clinical study. The window size of frame 6 (top row), frame 18 (middle row) and frame 28 (bottom row) is set to [0, 0.1], [0, 0.15] and [0, 0.8], respectively. Each image is shown in a transverse view. ‘A’ represents the artery ROI, 'L' represents the lung ROI and ‘T’ is the tumor ROI.}
    \label{fig:7}
\end{figure*}

\begin{figure*}[t]%[th!b]
    \centering
        \includegraphics[width=1.0\textwidth]{"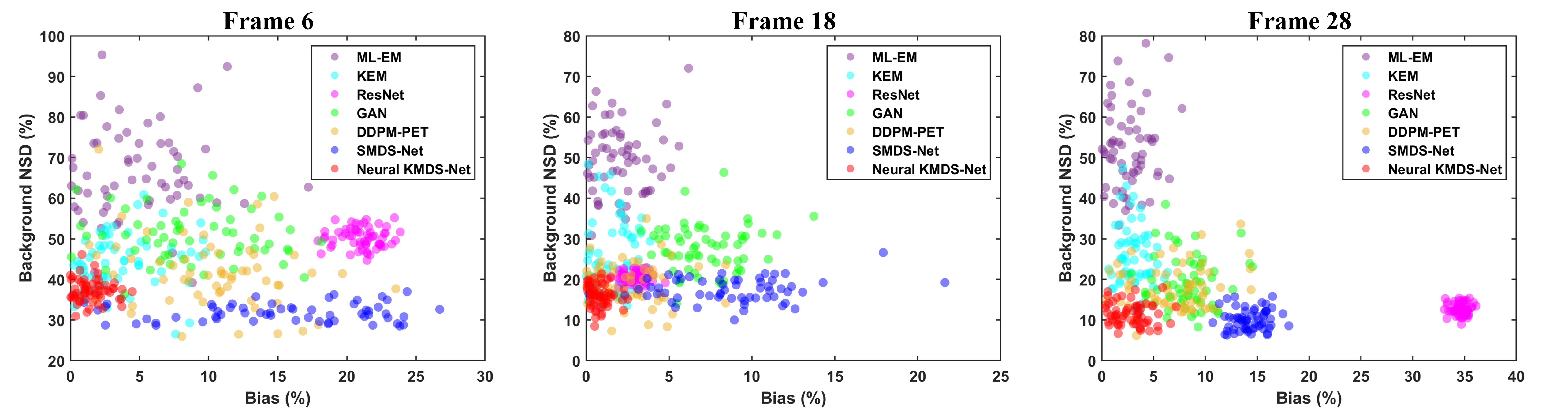"}
        \caption{Plots of bias of tumor (frame 6 and 28) and artery (frame 18) versus lung background NSD. Each point represents the evaluation result from a realization of the test reference image in Fig. \ref{fig:7}. The locations of the tumor ROI, artery ROI, and lung background ROI are illustrated in Fig. \ref{fig:7}.}
    \label{fig:8}
\end{figure*}

\begin{table*}[t]\scriptsize%[th!b]
    \caption{Quantitative comparison of the global chest image for frame 6, 18 and 28 using PSNR and SSIM in the clinical study. Mean is reported. Best results are marked in \textbf{BOLD}.} \label{tab6}
    \centering
    \begin{tabular}{ c l c c c c c c c}
        \hline\hline
        \multicolumn{2}{c}{Methods} & ML-EM & KEM & ResNet & GAN & DDPM-PET & SMDS-Net & Neural KMDS-Net \\
        \hline
        \multirow{3}{*}{PSNR} & Frame 6 & 25.795 & 28.702 & 25.580 & 30.826 & 31.908 & 32.403 & \textbf{33.081} \\
        & Frame 18 & 28.238 & 31.717 & 31.449 & 32.532 & 33.603 & 33.125 & \textbf{35.289} \\
        & Frame 28 & 33.297 & 35.237 & 32.512 & 37.164 & 37.460 & 37.748 & \textbf{40.795} \\
        \hline
        \multirow{3}{*}{SSIM} & Frame 6 & .645 & .741 & .746 & .822 & .847 & .883 & \textbf{.884} \\
        & Frame 18 & .735 & .843 & .867 & .869 & .897 & .927 & \textbf{.934} \\
        & Frame 28 & .811 & .870 & .898 & .924 & .928 & .948 & \textbf{.963} \\   
        \hline\hline
    \end{tabular}
\end{table*}

\section{Real patient study}
\label{sec:real}
\subsection{Data acquisitions}
We further employ neural KMDS-Net for dynamic PET denoising for real patient datasets. 65 patients scans are performed using the DigitMI 930 PET/CT system. The acquisition starts right after the injection of $1.85\sim 4.44$ MBq/kg $^{18}$F-FDG and lasts 60 minutes. Fifty-five patient data are used for training, while the remaining ten patient data are used for testing. Informed consent is obtained from all human subjects involved in this retrospective study. Approval of all ethical and experimental procedures and protocols is granted by the institutional review board of the Beijing Friendship Hospital, Capital Medical University, Beijing, China (approval number 2022-P2-314-01. date:10/08/2022). The one-hour data are divided into 28 time frames following the schedule \(12 \times 60\)s and \(16 \times 180\)s. The original image matrix size is $192\times 192\times 96$ per bed position. Considering the limitations of GPU memory, the original image is continuously cropped along the third dimension into small block images of size $192\times 192 \times 8$.

To generate the training data, the total acquired data is reconstructed as the label image using the vendor software. The acquired raw data is down-sampled to 1/10-th of the total counts and reconstructs as the input low-dose image. To increase the training data, 3 low-dose realizations from the high-dose data for each patient data are generated. For the test data, sixty noisy realizations for each patient data are simulated.

\subsection{Results of denoised PET images}
Fig. \ref{fig:7} shows the denoised chest PET images using different methods for frame 6, 18 and 28. All DL-based denoising models are trained exclusively using patients' chest PET data. The ML-EM has the highest noise and KEM effectively reduces image noise compared to ML-EM. ResNet and GAN successfully suppress the noise but introduce noticeable bias in the tumor and artery regions (Fig. \ref{fig:8}), which similarly occurs in the results produced by SMDS-Net. DDPM-PET outperforms the above methods but introduces some textural details that are not present in the reference images. In comparison, the images denoised using the proposed neural KMDS-Net shows significant improvement with lower bias and noise. These observations are consistent with the quantitative results shown in Fig. \ref{fig:8} and Table \ref{tab6}.

\begin{figure*}[t]%[th!b]
    \centering
        \includegraphics[width=1.0\textwidth]{"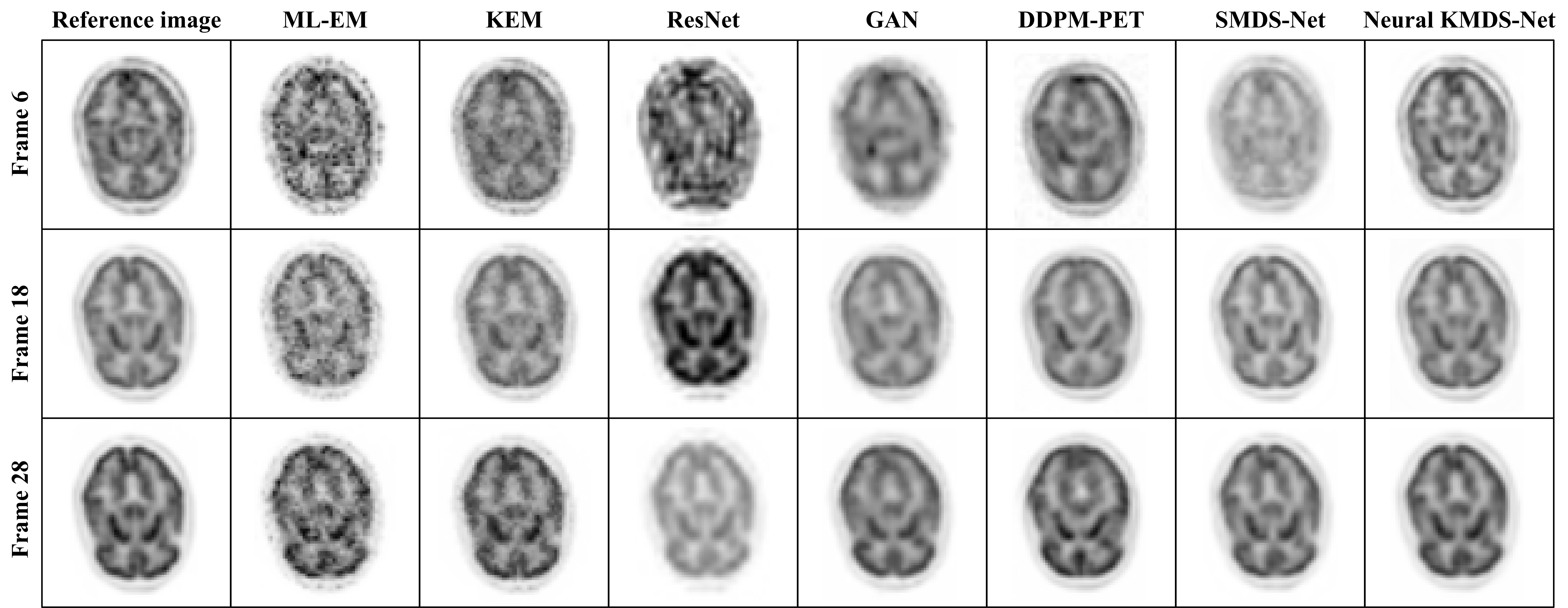"}
        \caption{True activity images and denoised images by different methods in a clinical brain study. The window size of frame 6 (top row), frame 18 (middle row) and frame 28 (bottom row) is set to [0, 0.04], [0, 0.2] and [0, 1], respectively. Each image is shown in a transverse view.}
    \label{fig:9}
\end{figure*}

To further evaluate the generalization of the data-driven denoising methods in clinical scenarios, we train the neural network models using whole-body PET data from multiple bed positions and validate them on brain data. Fig. \ref{fig:9} shows the denoised brain images using different methods for frame 6, 18 and 28. The ML-EM results suffer from heavy noise. The KEM suppresses noise well but still suffers from noise and artifacts due to insufficient kernel representation. ResNet produces severe bias and artifacts on all 3 frames. GAN, DDPM-PET and SMDS-Net produce poor results such as frame 6, failing to preserve some structural contrast and details. In comparison, the images by the proposed neural KMDS-Net demonstrate a significant improvement with clearest structures and lowest noise. The quantitative results are shown in Table \ref{tab7}. Our method achieves the highest PSNR and SSIM, demonstrating that our proposed neural network has strong generalization capabilities compared to other deep learning methods.

\begin{table*}[t]\scriptsize%[th!b]
    \caption{Quantitative comparison of the global brain image for frame 6, 18 and 28 using PSNR and SSIM in the clinical study. Mean is reported. Best results are marked in \textbf{BOLD}.} \label{tab7}
    \centering
    \begin{tabular}{ c l c c c c c c c}
        \hline\hline
        \multicolumn{2}{c}{Methods} & ML-EM & KEM & ResNet & GAN & DDPM-PET & SMDS-Net & Neural KMDS-Net \\
        \hline
        & Frame 6 & 28.564 & 34.267 & 18.720 & 30.760 & 31.405 & 27.910 & \textbf{35.257} \\
        PSNR & Frame 18 & 30.578 & 36.391 & 23.275 & 34.062 & 31.350 & 34.886 & \textbf{37.985}  \\
        & Frame 28 & 31.348 & 35.994  & 26.054 & 34.626 & 31.879 & 36.787 & \textbf{38.788} \\
        \hline
        & Frame 6 & .730 & .883 & .565 & .787 & .795 & .716 & \textbf{.912} \\
        SSIM & Frame 18 & .839 & .946 & .898 & .893 & .830 & .968 & \textbf{.971} \\
        & Frame 28 & .859 & .944 & .934 & .915 & .859 & .968 & \textbf{.969} \\
        \hline\hline
    \end{tabular}
\end{table*}

\section{Discussion}
\label{sec:dis}
This paper develops a neural KMDS-Net algorithm for dynamic PET image denoising.
The multidimensional sparse representation of the tensor is used to establish the spatial-temporal correlation of the 4D coefficient image in the kernel space for dynamic PET. A neural network is then utilized to facilitate adaptive parameter optimization.
Compared with other DL-based methods, our network is a lightweight ($\sim$1 M) model with strong denoising capability for dynamic PET. 

The tradeoff of bias and noise is challenging for dynamic PET image denoising. The model-based methods usually produce results with low bias and high noise. The DL-based methods, on the other hand, suppress noise significantly yet may introduce significant bias. Using deep learning modules in a physical model-driven solution may achieve a better tradeoff, but the denoising results may still be undesirable. Fully embedding deep learning priors into physical models may improve this issue. Therefore, in this paper, we construct a KMDS model in the image domain and substitute each module with a neural network, forming an end-to-end trainable neural network. Simulation and real results demonstrate that our approach produces promising results.

The main differences between our proposed method and SMDS-Net are as follows. I) The sparse coding space in which $\boldsymbol{\beta}$ sits.
SMDS-Net uses a low-dimensional subspace to exploit the low-rankness, while we employ an isometric kernel space to establish spatial-temporal correlations for dynamic PET. It has been shown that the kernel method effectively leverages the prior knowledge of different frames in dynamic PET and may performs better in image denoising.
II) The learnable parameters from (9) to (11). SMDS-Net uses dictionaries to explore the features and structures of tensors in different dimensions. Our method uses convolutional kernels with stronger representation ability to optimize $\boldsymbol{\beta}$. The sparse coding problem of $\boldsymbol{\beta}$ can be learned more accurately by neural networks.

It should be noted that the proposed method achieves slightly inferior quantitative results at the first frame relative to several reference methods. Although the exact reason underlying this phenomenon remains under further investigation, we conjecture that it may be related to the relatively restricted model capacity, which might be insufficient to comprehensively characterize the diverse data distributions throughout the whole dynamic PET series, particularly for the early frame with low-count statistics.
In future research, we intend to enhance the model expressiveness by introducing advanced architectural units such as Mamba and Transformer, or enlarging the channel number of convolutional layers to scale up model capacity, so as to achieve stable and promising denoising performance over all temporal frames.

Table \ref{tab4} reveals that the performance of neural KMDS-Net is improved by stacking MDSB modules 20 times. However in the kernel representation module, stacking just two convolutional layers yields the optimal performance of neural KMDS-Net (as shown in Table \ref{tab3}). This illustrates the importance of physical model priors when designing neural network structures. We are considering integrating compressive sensing into the physical model, combining it with kernel methods and multidimensional sparse coding to design a better neural network architecture.

\st{In the experiments with clinical data, it has been observed that the comparison methods like ResNet, GAN, DDPM-PET and SMDS-Net demonstrate good performance on later frames yet are quite poor on the earlier ones, showing severe artifacts or even distortion in the denoised PET images. This could be attributed to the fact that all frames are inputted into the network together, with short frames contributing less to the network's total loss due to their low intensity values. In contrast, by implicitly integrating physical priors into the neural network, our proposed neural network has shown much better performance in the short frames, which is quite important for observing the tracer distribution close to the injection time.}

On the clinical dataset, the proposed Neural KMDS-Net shows performance degradation on frames earlier than frame 6, which is consistent with the observation in the initial frame of the simulation study. This limitation can be explained from two perspectives. On one hand, the relatively modest model capacity may restrict its ability to adapt to the wide and diverse data distribution across the full dynamic PET series. On the other hand, very early frames suffer from extremely low tracer uptake, leading to barely visible anatomical structures and organ boundaries. Since the proposed method leverages inter-frame structural consistency, the absence of reliable structural information in these early frames leads to degraded denoising performance. Once clear anatomical structures become visible starting from frame 6, our method achieves superior performance over all competing methods.
In future research, we will further investigate dedicated mechanisms to handle extremely early frames with indistinct anatomical structures, so as to improve the overall temporal stability and clinical utility of the proposed approach.

In our generalization experiments, we utilize the network trained exclusively on dynamic PET whole-body data and test on brain dynamic PET sequences. The physical model-dependent approach such as KEM can effectively preserve image details but include significant noise. The deep learning methods, such as ResNet, GAN and DDPM-PET, control noise well but tend to oversmooth the image details. In contrast, by designing neural networks based on physical models, our proposed neural KMDS-Net combines the high generalizability of the model-based methods with the strong denoising capabilities of DL-based methods. Incorporating more physical priors to guide the design of neural networks may further enhance the generalization performance of our model. In addition, we find that the neural network achieves the best performance when the acquisition time of test frame is similar to that of training frame. The network’s performance degrades when the acquisition time of the training and test data does not match. Therefore, to achieve optimal performance using the proposed neural KMDS-Net, a new refined training procedure may be needed when the noise levels of the test data differ from those of the training data.

The short frames in dynamic PET images are usually quite noisy due to the short frame durations, which makes denoising more difficult. In this paper, we use kernel methods and multidimensional sparse tensors to establish in-depth relationship between temporal frames. Our future work will focus on using more advanced neural network structures in each optimization module to achieve better denoising results.

\section{Conclusion}
\label{sec:con}
% In this paper, a novel interpretable neural network for dynamic PET image denoising is proposed. We develop a kernel space-based multidimensional sparse model that captures the spatial correlation and structural consistency between temporal PET frames, and construct an end-to-end deep neural network to enable adaptive parameter optimization. Computer simulation and patient studies show that the neural KMDS-Net achieves superior performance over existing methods in dynamic PET denoising. The proposed method may be used as an effective solution for image denoising in dynamic PET.
This paper proposes an end-to-end convolutional neural network designed based on a physical model to improve the quantitative accuracy of low-dose dynamic PET frames. In the methodology section, we detail the design principles for each module of the neural network architecture. In the experimental section, we determine the depth of network architecture and the size of network features through experimental studies. Moreover, all comparison methods are evaluated both qualitatively and quantitatively on simulated and clinical data. The proposed method demonstrates superior performance compared to all the benchmark approaches, particularly in the recovery of low-dose frames.

Benefiting from the neural network architecture guided by the physical model, our method enables the rapid acquisition of high-quality dynamic PET images in clinical scenarios. Furthermore, the application of our method may be extended to various imaging scenarios such as hyperspectral imaging. However, despite its promising performance, the proposed method still has potential for further enhancement.
(1) Convolutional layers are used to replace several steps in the optimization process of the denoising model. Future work will include using more advanced network mechanisms such as spatial-temporal joint attention. (2) Only the kernel method and multidimensional sparse coding are integrated in designing the neural network. The denoising performance may be further improved by incorporating more explainable strategies such as compressed sensing. (3) The optimal network parameters are numerically determined through experiments. In future work, they may be automatically adjusted in a more learnable manner. (4) In clinical scenarios, the performance of the algorithm may be affected when the noise distribution changes significantly, which needs to be explored in our future studies.

\section*{Acknowledgements}
This work was supported in part by the National Natural Science Foundation of China under Grant 62571476 and Grant 62401516.

\section*{References}
\bibliographystyle{unsrt}
\bibliography{refs}

\end{document}